\documentclass[sigconf, preprint]{acmart}

\AtBeginDocument{%
  \providecommand\BibTeX{{%
    \normalfont B\kern-0.5em{\scshape i\kern-0.25em b}\kern-0.8em\TeX}}}

\setcopyright{acmcopyright}
\copyrightyear{2020}
\acmYear{2020}
\acmDOI{}

\acmConference[arXiv]{arXiv}{May 17}{2020}
\acmBooktitle{arXiv}
\acmPrice{15.00}
\acmISBN{978-1-4503-XXXX-X/18/06}

\usepackage{xcolor}
\usepackage[normalem]{ulem}
\definecolor{dgreen}{rgb}{0,.6,0}

\begin{document}

\title{Separation of Memory and Processing in Dual Recurrent Neural Networks}

\author{Christian Oliva}
\affiliation{%
  \institution{Escuela Polit\'{e}cnica Superior,\\
  Universidad Aut\'{o}noma de Madrid}
  \state{28049 Madrid, Spain}
}
\email{christian.oliva@estudiante.uam.es}

\author{Luis F. Lago-Fern\'{a}ndez}
\affiliation{%
  \institution{Escuela Polit\'{e}cnica Superior,\\
  Universidad Aut\'{o}noma de Madrid}
  \state{28049 Madrid, Spain}
}
\email{luis.lago@uam.es}

\renewcommand{\shortauthors}{Oliva \& Lago-Fern\'{a}ndez}

\begin{abstract}
We explore a neural network architecture that stacks a recurrent layer and a feedforward layer that is also connected to the input, and compare it to standard Elman and LSTM architectures in terms of accuracy and interpretability. When noise is introduced into the activation function of the recurrent units, these neurons are forced into a binary activation regime that makes the networks behave much as finite automata. The resulting models are simpler, easier to interpret and get higher accuracy on different sample problems, including the recognition of regular languages, the computation of additions in different bases and the generation of arithmetic expressions.
\end{abstract}






\maketitle

\section{Introduction}

Machine learning techniques, and more specifically deep neural networks (DNNs), have become essential for a wide range of applications, such as image classification \cite{Litjens201760, 7803544}, speech recognition \cite{DBLP:journals/corr/abs-1303-5778} or natural language processing \cite{mikolov_2010, DBLP:journals/corr/SutskeverVL14, DBLP:journals/corr/ChoMGBSB14}. Notwithstanding, nowadays these deep models are still not dominant in many applications due to the common belief that simpler approaches, such as linear models or decision trees, provide better interpretability. Hence techniques for interpreting DNNs are becoming popular in fields like image classification \cite{MONTAVON20181} and sequence modeling including music composition and natural language generation \cite{DBLP:journals/corr/KarpathyJL15, DBLP:journals/corr/StrobeltGHPR16}. However, and in spite of all the effort, the interpretation and understanding of DNNs is still an open question that deserves further research.

Recurrent Neural Networks (RNNs) are a kind of deep network aimed at sequence modeling, where the depth comes from a recurrent loop in the network architecture that forces a backpropagation through several time steps when the gradients are computed \cite{graves_book}. Since their introduction, RNNs have been shown to be Turing equivalent \cite{SIEGELMANN1995132}, and many authors have studied the ability of these networks to model different kinds of formal languages \cite{DBLP:journals/neco/ZengGS93, omlin_and_giles_1996, gers_and_schmidhuber_2001}. The interpretability of RNNs has been often addressed by quantization approaches that try to reduce the network to a set of rules ({\em rule extraction}) \cite{omlin_and_giles_1996, jacobsson_05, DBLP:journals/corr/abs-1709-10380}, usually in the form of a deterministic automaton \cite{DBLP:conf/nips/GilesMCSCL91, casey_1998, cohen_2017}. 

More recently, the picture has been completed with the introduction of Memory Augmented Neural Networks (MANNs) \cite{DBLP:journals/corr/abs-1807-08518}, where a standard, and usually recurrent, neural network is enhanced with some type of external memory \cite{DBLP:journals/corr/GravesWD14, hybrid_ntm, Bernardy2018CanRN}. Results of these new models on complex problems seem very promising. Additionally, as much of the computational power of these networks relies on the memory, the complexity of the neural component is reduced, hence also improving the overall model interpretability. The connection with general automata seems obvious in this case, with the neural network implementing a kind of finite state neural processor and different memory schemes leading to different types of abstract models, from pushdown automata to complete Turing machines.

Following these ideas, in this article we explore a neural network architecture that combines a simple feedforward processing layer with a recurrent layer that implements a sort of memory. When this \textit{Dual} network is trained to process temporal sequences, the recurrent layer is used to keep track of only the essential information that must be preserved along time, without performing any additional computation. The feedforward layer combines in turn the input and the memory content to provide the final network's output. This separation of roles seems to be beneficial for learning, since much of the computational power is discharged from the recurrent connection, and at the same time improves interpretability. From a more abstract point of view, we find that the network may be reduced to a Mealy machine, just as a simple recurrent network can be reduced to a finite automaton in the form of a Moore machine \cite{DBLP:conf/icann/OlivaL19}. Mealy machines are simpler than Moore machines with respect to the number of states, and so the models trained using this architecture are also simpler and more easily interpretable. We study the network's capacity to solve several simple problems, including the recognition of regular languages, the computation of additions in different numerical bases and the generation of arithmetic expressions, and compare the results to standard RNNs with Elman architecture \cite{elman_1990} and Long Short Term Memory (LSTM) networks \cite{lstm}. We show that the Dual architecture can enhance the prediction accuracy and at the same time generate highly interpretable models. 
 
The article is organized as follows. In section \ref{sec:models} we introduce the different network architectures used in all our experiments, including the Elman and LSTM networks that are used as a benchmark for comparison. In section \ref{sec:experiments} we describe the data and the experiments. Section \ref{sec:results} presents and analyzes the results. Finally, in section \ref{sec:conclusion} we present the conclusions and discuss future lines of research.

\section{Networks}
\label{sec:models}

\subsection{Elman RNN with noisy recurrence}
\label{subsec:elmanRNN}

The Elman RNN \cite{elman_1990} is the simplest neural network architecture where recurrence is introduced. It adds a time dependence to the internal layer, making the activity in this layer depend on its output for the previous time step. Here we use a modified Elman RNN with one single hidden layer based on \cite{DBLP:conf/icann/OlivaL19}, where noise is introduced in the activation function of the recurrent layer units. The network behavior is governed by the following equations:

\begin{flalign}
\label{eq::noise_rnn_h}
& h_{t} = \tanh(W_{xh}x_{t} + W_{hh}h_{t-1} + X_{\nu} \circ h_{t-1} + b_{h}) \\
\label{eq::elman_rnn_y}
& y_{t} = \sigma(W_{hy}h_{t} + b_{y})
\end{flalign}

\noindent where $h_{t}$ and $y_{t}$ represent the activation of the hidden and output layers, respectively, at time $t$, and $x_{t}$ is the network input. The model depends on weight matrices $W_{xh}$, $W_{hh}$ and $W_{hy}$, and bias vectors $b_{h}$ and $b_{y}$. In particular $W_{hh}$ represents the weights in the recurrent connection that makes a explicit dependence of $h_{t}$ on $h_{t-1}$. The recurrent connection also includes the noisy term $X_{\nu} \circ h_{t-1}$, where $X_{\nu}$ is a random vector whose elements are drawn from a normal distribution with mean $0$ and standard deviation $\nu$ each time the value of $h_{t}$ needs to be computed. The $\circ$ operator denotes an element-wise product. This noise is more effective for very active neurons, being negligible for silent ones. This way, the effect of the noise on the overall network's behavior is to force the operation of neurons in an almost binary fashion \cite{DBLP:conf/icann/OlivaL19}. 

\begin{figure}[h]
	\centering
	\includegraphics[width=0.35\textwidth]{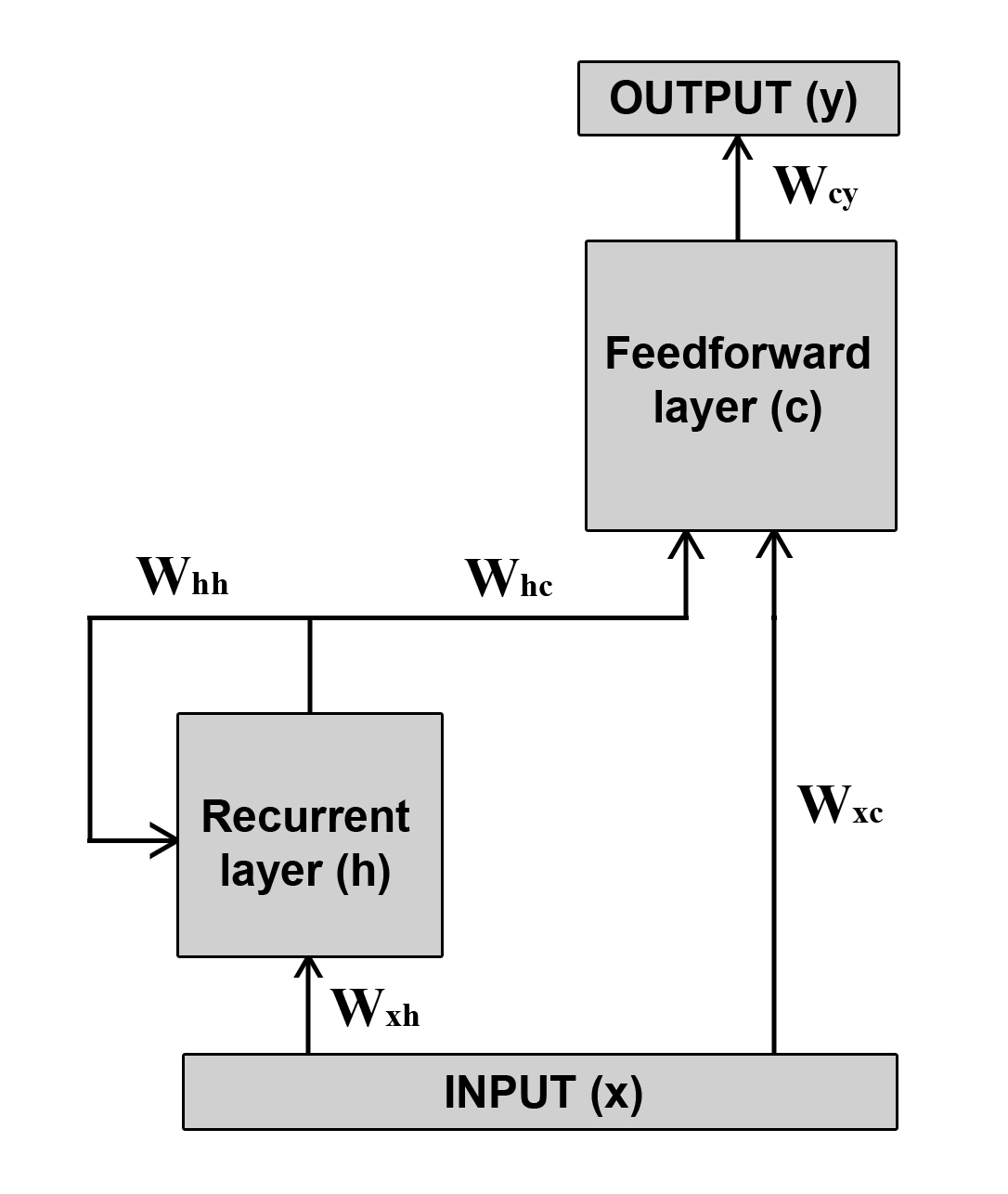} \\
	\caption{Connection diagram of the Dual RNN}
	\label{fig:dual_diagram}
\end{figure}

\subsection{Dual RNN}
\label{subsec:dualRNN}

The second network architecture that we explore combines the noisy recurrent layer described in section \ref{subsec:elmanRNN} with a feedforward layer that receives its input from both the recurrent and the input layers. We include a final sigmoid layer on top to provide the network output (see figure \ref{fig:dual_diagram}). The network activity is controlled by the following set of equations:

\begin{flalign}
\label{eq::dual_rnn_h}
& h_{t} = \tanh(W_{xh}x_{t} + W_{hh}h_{t-1} + X_{\nu} \circ h_{t-1} + b_{h}) \\
\label{eq::dual_rnn_c}
& c_{t} = \tanh(W_{xc}x_{t} + W_{hc}h_{t} + b_{c}) \\
\label{eq::dual_rnn_y}
& y_{t} = \sigma(W_{cy}c_{t} + b_{y}) 
\end{flalign}

\noindent where the symbols have the same interpretation as before. The idea behind this implementation is to permit the recurrent units focus on just processing the information that must be maintained along time, and discharge it from other processing that can be performed by a simple feedforward layer. This way, we expect that the recurrent layer acts as a memory that learns to deal with time dependencies, while the feedforward layer learns to combine the input and the memory to provide an answer given the current input. 
We call this network a \textit{Dual} RNN. 

\subsection{LSTM network}
\label{subsec:lstmNetwork}

Finally, we consider LSTM networks \cite{lstm}, which are one of most commonly used RNN models, and a usual benchmark in many sequence modeling applications. They introduce memory cells and gate units to build an architecture that keeps an almost constant error signal along time, avoiding the vanishing and exploding gradient problems. Here we use the standard Keras implementation \cite{lstm_keras}.

\subsection{Network training}
\label{subsec:networkTraining}

All the networks above are trained to minimize a cross-entropy loss with L1-regularization using a standard gradient descent optimizer. Regularization is applied only to the weight matrices, but not to the biases. In the Dual RNN only the weights in the recurrent layer ($W_{xh}$, $W_{hh}$ and $W_{hc}$) are regularized.

\section{Experiments and Data}
\label{sec:experiments}

We test our network on three different problems: the recognition of regular grammars, the addition in different numerical bases and the generation of arithmetic expressions. In the following we provide a detailed description of the data and the experiments. 

\subsection{Recognition of regular languages}
\label{subsec:experiments_tomita}

The first test considers the ability of the networks to recognize the \textit{Tomita Grammars} \cite{tomita}, seven regular languages defined on the alphabet \{$a$, $b$\} that are a standard in rule extraction \cite{DBLP:journals/corr/abs-1709-10380}. A full description of the seven Tomita grammars is provided in table \ref{Tab::tomitaGramars}. We use the experiments and results described in  \cite{DBLP:conf/icann/OlivaL19} as a reference benchmark. The datasets for training and evaluating the networks are generated as therein described, including the $\$$ symbol used as string separator. Networks are evaluated by measuring their recognition accuracy on the test set.

\begin{table}
	\begin{center}
		\caption{Description of the $7$ Tomita Grammars}
		\begin{tabular}{ll}
			\toprule
			\textbf{Name} & \textbf{Regular language}  \\
			\midrule
			Tomita1 & Strings with only $a$'s. \\
		    Tomita2 & Strings with only sequences of $ab$'s. \\
			Tomita3 & Strings with no odd number of \\
			        & consecutive $b$'s after an odd number \\
			        & of consecutive $a$'s. \\
			Tomita4 & Strings with fewer than 3 \\
			        & consecutive $b$'s. \\
			Tomita5 & Strings with even length with an \\
			        & even number of $a$'s. \\
		    Tomita6 & Strings where the difference \\
			        & between the number of $a$'s and $b$'s is a \\
			        & multiple of 3. \\
			Tomita7 & $b^{*}a^{*}b^{*}a^{*}$ \\
			\bottomrule
		\end{tabular}
		\label{Tab::tomitaGramars}
	\end{center}
\end{table}

\subsection{Addition}
\label{subsec:experiments_addition}

The next problem consists of predicting the result of a sum of two numbers that are presented to the network digit by digit. Different numerical bases are considered. In base $B$, the network inputs at a given time step are two numbers between $0$ and $B-1$. The network is expected to return a single number in the same range which is the result (without carry) of the addition of the two inputs plus the possible carry generated in the previous time step. One example of input and output for $B=2$ is shown in table \ref{tab:sum_example}, where the strings must be read from left to right, contrary to the usual binary number representation. Note that we also include the separator symbol $\$$ in the input strings, which represents the end of a particular sum. The $\$$ symbol appears always in both input strings at the same time, and the expected output in this case is just the carry after the last two digits have been added.    

We consider two different randomly generated datasets, each consisting of two input strings of length $200000$ and the corresponding output string of the same length. The generation probability for the $\$$ symbol is set to 0.1, and the rest of symbols are all equally probable. The first dataset is used to train the networks, the second is used for test. As before, the networks are evaluated by measuring their prediction accuracy on the test set.

Although the grammar associated to this problem is regular as in the previous case, the problem complexity increases with $B$ and the interpretation of a standard RNN solution becomes nontrivial. As we show in the results section, this kind of problem illustrates the benefits of using a network that separates the recurrent memory from the main processing path, such as the Dual RNN. Since the only information that needs to be remembered is the carry, the discharge of computing power in the recurrent layer represents a great advantage in this problem.

\begin{table}
    \centering
    \caption{Example of inputs and output for the addition problem in base 2.}
    \begin{tabular}{ll}
        \toprule
        Input 1 & $\$0101110101011010\$01011101$ \\
        Input 2 & $\$1101010110101011\$01011010$ \\
        Output  & $01010011000001100100101000$ \\
        \bottomrule
    \end{tabular}
    \label{tab:sum_example}
\end{table}

\subsection{Generation of arithmetic expressions}
\label{subsec:experiments_expressions}

Finally we test the models on a generation task. The networks are trained to predict the next symbol in an arithmetic expression that includes operators ($*$, $/$, $+$, $-$), parentheses and operands represented by the single symbol $a$, as well as the string separator $\$$. Both the training and test datasets are single strings with $200000$ symbols randomly generated according to the grammar rules: \\

\noindent $S$ $\rightarrow$ $S$ $op$ $T$ $|$ $T$ \\
$T$ $\rightarrow$ $a$ $|$ $($ $S$ $)$ \\
$op$ $\rightarrow$ $+$ $|$ $-$ $|$ $*$ $|$ $/$ \\

\noindent with the maximum parentheses depth limited to $5$. Table \ref{Tab::examples_arithmetic} shows the first characters of the training dataset. Note that any substring within $\$$ symbols is a valid arithmetic expression with parentheses depth not greater than $5$.

\begin{table}
	\begin{center}
		\caption{First 156 input characters from the training dataset in the arithmetic expression generation problem.}
		\hrule
		\vskip 0.1in
        \begin{verbatim}
$((a)/a/a)+a+(a-a-a+((((a)))/(((a))+a/a)))$(a/(a-a-((
(a-a*a+a))/a)-((a/(a)/(a/a/a)))+a))$a-(((a/a*a)+a)-a)
-(a*(a)/(a))$a$(((((a)-(a-a+a-a)/a/a+a)/((a*a/a)/(...
        \end{verbatim}
        \vskip -0.05in
        \hrule
		\label{Tab::examples_arithmetic}
	\end{center}
\end{table}

In this problem we evaluate the trained networks by measuring their ability to generate correct arithmetic expressions. Starting with an initial $\$$ symbol, the networks are asked to generate a string with the next $50000$ symbols. This string is then split into substrings delimited by $\$$ and the grammatical correction of these substrings is checked. We define the accuracy of a network as the percentage of correctly generated expressions. Table \ref{tab::sample_strings} shows some examples of correct and incorrect strings generated by networks trained on this problem. 

\begin{table}
	\centering
	\caption{Some examples of correct and incorrect strings generated by networks trained on the arithmetic expressions problem.
	}
	\label{tab::sample_strings}
	\begin{tabular}{lcc }
		\toprule
		\textbf{Example} & \textbf{Test} & \textbf{Why?} \\
		\midrule
		$a-((a)+a)*/a+(a)$ & $\times$ & $*/$ \\
		$(a+())-a*(((a)))$ & $\times$ & $()$ \\
		$a-((a))+a)/a+(a)$ & $\times$ & Depth -1 \\
		$((a(a))+a*a/((a))$ & $\times$ & Depth +1 \\
		$((a(a))+a*a/((a)))$ & $\surd$ & \\
		\bottomrule
	\end{tabular}
\end{table}

\section{Results}
\label{sec:results}

\subsection{Tomita Grammars}
\label{subsec:results-tomita}

For the recognition of Tomita grammars only the noisy Elman and the Dual networks are considered. In both cases it is not difficult to find network parameters that allow for a perfect classification with a $100\%$ accuracy in all tests. For the noisy Elman network we use $10$ units in the recurrent layer, learning rate $r = 0.01$, L1 regularization $\lambda = 0.1$, a batch size of $10$ and an unfold length of $25$. The Dual network uses the same parameters, with   $10$ additional units in the feedforward layer. The networks are trained for $1000$ epochs and we use an adaptive noise level that starts at $\nu = 0.0$ and linearly increases to reach a maximum value of $\nu = 1.0$ at epoch $500$. We do not need to include the \textit{shocking} mechanism reported in \cite{DBLP:conf/icann/OlivaL19}, maybe due to the use of adaptive noise. 

\begin{figure}[h]
    \centering
    \includegraphics[width=0.2\textwidth]{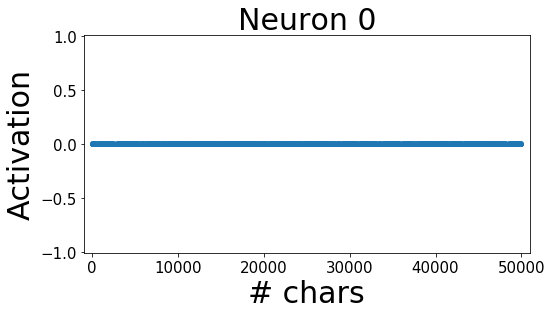}
    \includegraphics[width=0.2\textwidth]{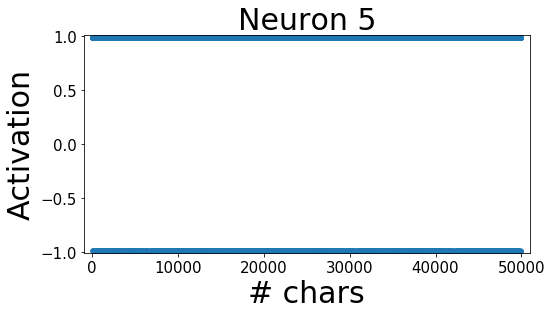}
    \includegraphics[width=0.2\textwidth]{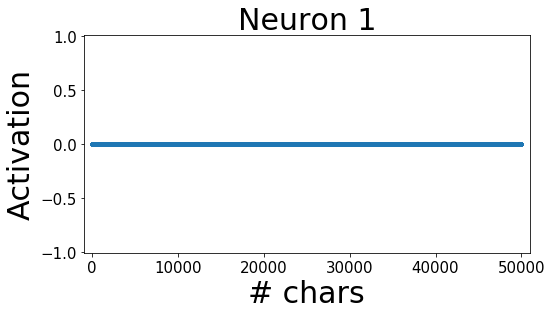}
    \includegraphics[width=0.2\textwidth]{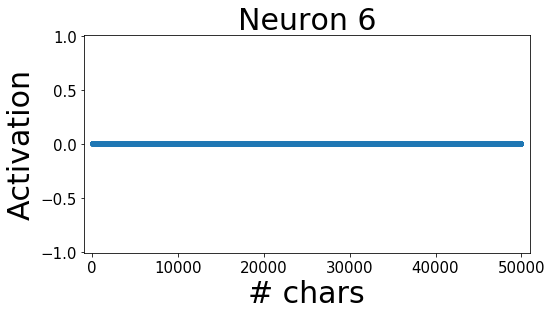}
    \includegraphics[width=0.2\textwidth]{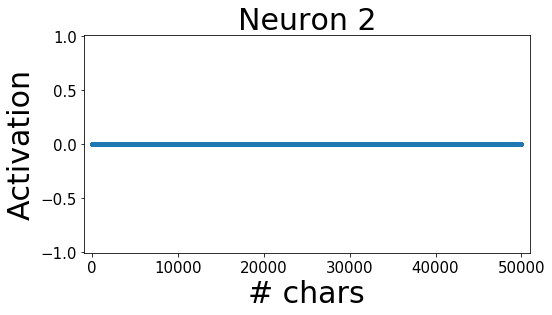}
    \includegraphics[width=0.2\textwidth]{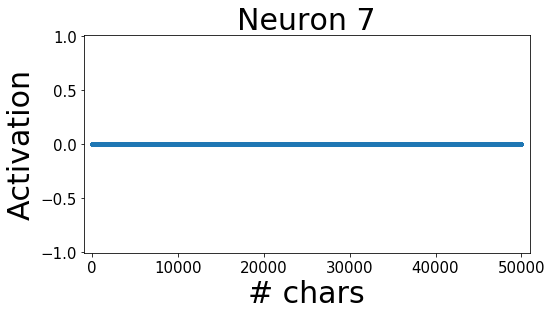}
    \includegraphics[width=0.2\textwidth]{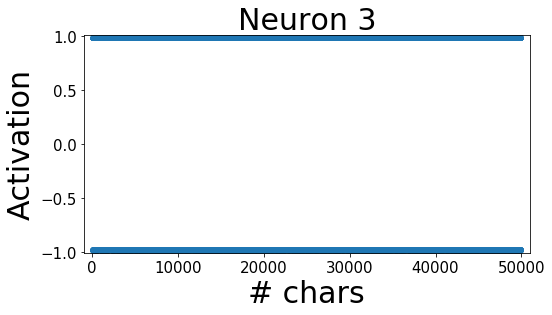}
    \includegraphics[width=0.2\textwidth]{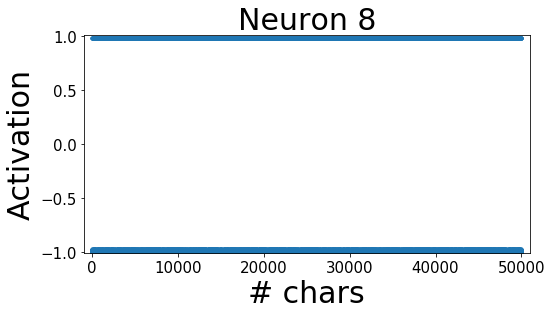}
    \includegraphics[width=0.2\textwidth]{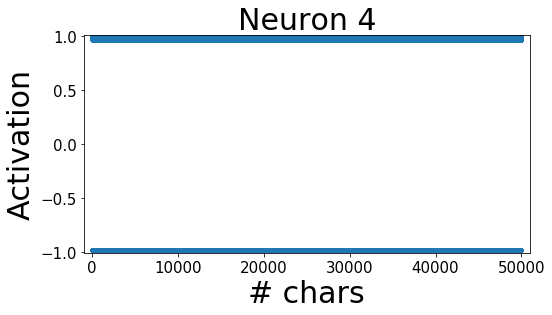}
    \includegraphics[width=0.2\textwidth]{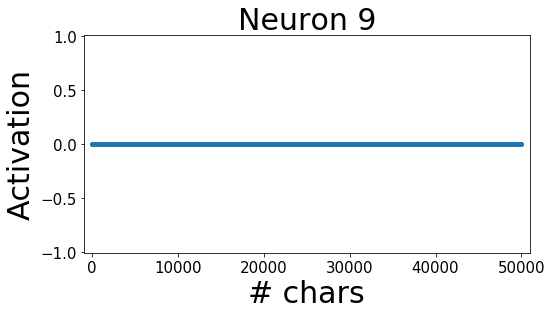}
    \caption{Activation of the recurrent layer neurons of a Dual RNN trained on the Tomita 6 problem in response to the 50000 symbols of a test string. Only neurons 3, 4, 5 and 8 get activated, and their output is always $+1$ or $-1$.}
    \label{fig:tomita6_activation}
\end{figure}

After training, the recurrent layer is strongly regularized, with only a few neurons participating in the generation of the network's output. Additionally, the active neurons operate in a binary regime, with their activity being pushed towards $+1$ or $-1$. No intermediate values are observed in the activation of recurrent units (see fig. \ref{fig:tomita6_activation}). Hence the activation patterns in the recurrent layer may be interpreted as a finite set of states, and state transitions in response to input symbols can be used to define a deterministic finite automaton (DFA) that summarizes the network behavior. This observation is general for all the networks that provide satisfactory test results, independently of their architecture.

\begin{figure}[h]
    \centering
    \includegraphics[width=0.3\textwidth]{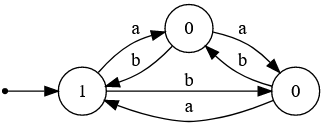}\\
    \includegraphics[width=0.3\textwidth]{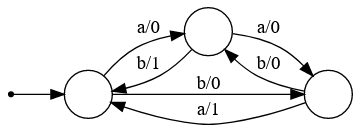}
    \caption{Minimum DFA extracted from an Elman RNN (top) and a Dual RNN (bottom), both trained on the Tomita 6 problem.}
    \label{fig:tomita6_elman}
\end{figure}

It is however very interesting to compare the automata obtained for the Elman and the Dual architectures. Figure \ref{fig:tomita6_elman} shows these automata for the Tomita 6 problem\footnote{Experiments performed on the other Tomita grammars provide similar results. These can be found in our GitHub repository: Anonymous, the link will be added after the review process.} after the application of a DFA minimization algorithm \cite{10.5555/1196416}. In the noisy Elman case the network's output depends only on the recurrent layer state. This is represented by associating an output symbol to each automaton state. This way, the noisy Elman network accepts an interpretation as a Moore machine (figure \ref{fig:tomita6_elman}, top). On the other hand, the Dual network's output depends both on the recurrent layer state and the input. We may incorporate this information into the corresponding DFA by adding the output symbol to the transition labels. In this case the resulting automaton is a Mealy machine (figure \ref{fig:tomita6_elman}, bottom).

Although the two automata seem in principle pretty similar, this observation is just valid for this particular problem. Note that for each state there exists only one valid transition with one single input symbol. Hence the Moore and Mealy machines have the same transitions graph. This panorama changes as soon as we consider more complex problems where the separation between input and memory represents a clear benefit in terms of interpretability. In those cases a network that admits an interpretation as a Mealy machine will be much simpler to understand. The addition problem considered next is one of such cases.

\subsection{Addition}
\label{subsec:results-addition}

\begin{figure}[h]
    \centering
    \includegraphics[width=0.15\textwidth]{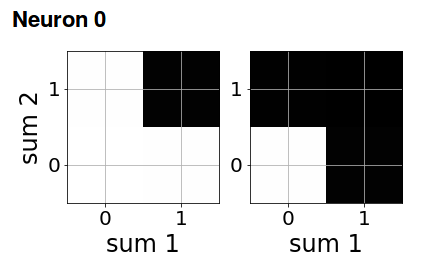}
    \includegraphics[width=0.15\textwidth]{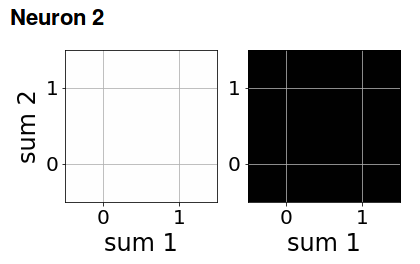}
    \includegraphics[width=0.15\textwidth]{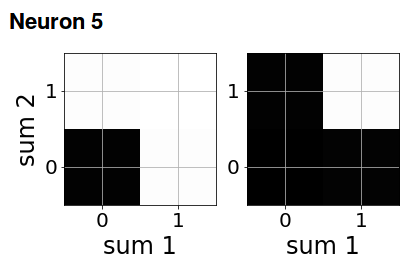}
    \caption{Activation plot of the three non-regularized neurons in an Elman RNN trained on the addition problem in base 2. For each neuron, the left plot represents the activation when there is no carry, while the right plot represents the activation when there is a carry from the previous step.}
    \label{fig:elman_sum2}
\end{figure}

\begin{figure}[h]
    \centering
    \includegraphics[width=0.35\textwidth]{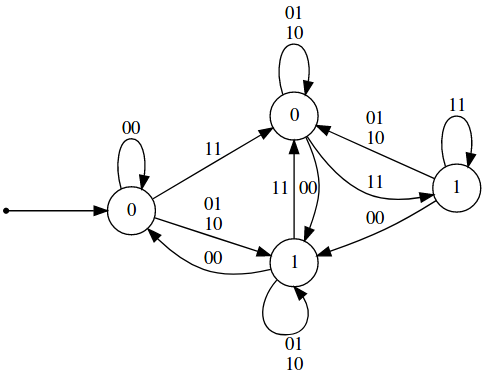}
    \caption{Minimum DFA extracted from an Elman RNN trained on the addition problem in base 2.}
    \label{fig:suma_moore}
\end{figure}

The addition problem described in section \ref{subsec:experiments_addition} has been approached using the noisy Elman and the Dual networks. The Elman RNN uses $20$ units in the recurrent layer. The rest of parameters are as in section \ref{subsec:results-tomita}. As before, after proper training the network is able to correctly predict all the samples in the test set. Regularization and binarization are also observed, with only a few recurrent units coding the solution in the form of a finite set of states. Figure \ref{fig:elman_sum2} shows the internal state space of one of such networks trained on the addition problem for $B=2$. Only the three active units are shown (the rest are always silent). And the three of them have a clear interpretation: the first neuron (N0) is learning the carry; the second (N2) keeps track of the carry in the previous time step; and the last one (N5) is dealing with the non-linearity of the binary addition problem, behaving as a XOR gate. A DFA can be extracted as before, and the network accepts again an interpretation as a Moore machine (see figure \ref{fig:suma_moore}).

From a theoretical point of view, only the state represented by the first two neurons must be kept in memory to solve the addition problem, since all the additional information needed to compute the state of N5 is contained in the current input. However, as the network has one single recurrent layer concentrating all the processing power, some recurrent units are forced to learn information that does not explicitly depend on the past history. For more complex problems this could imply an unnecessary waste of memory resources, also hindering the interpretability.

\begin{figure}[h]
    \centering
    \includegraphics[width=0.23\textwidth]{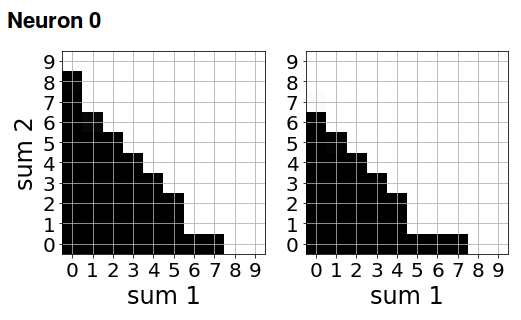}
    \includegraphics[width=0.23\textwidth]{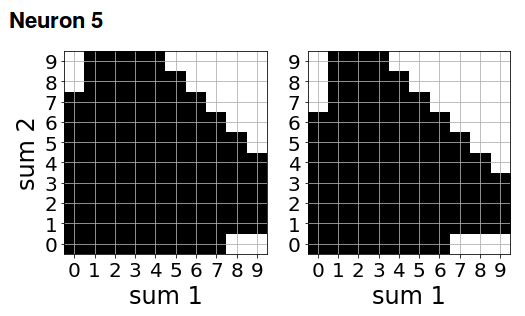}\\
    \includegraphics[width=0.23\textwidth]{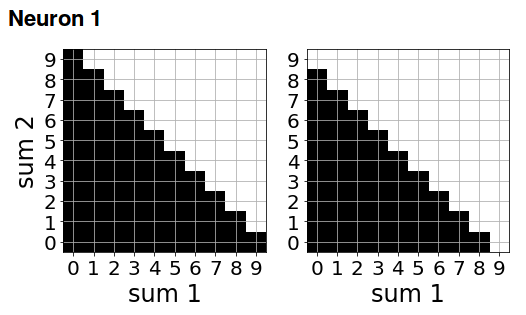}
    \includegraphics[width=0.23\textwidth]{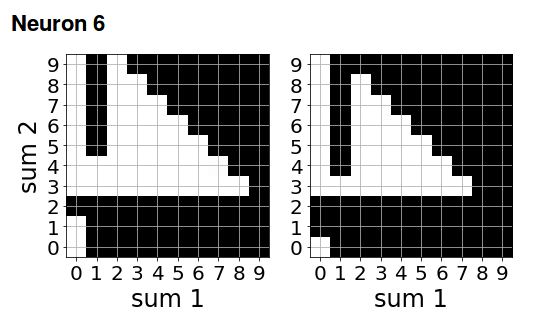}\\
    \includegraphics[width=0.23\textwidth]{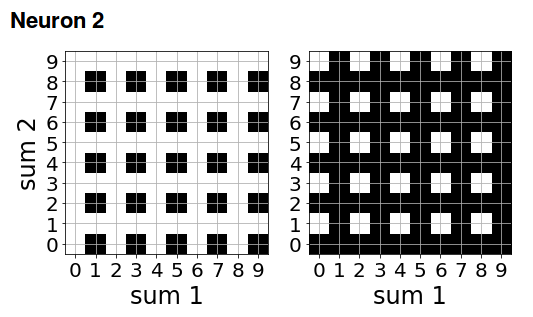}
    \includegraphics[width=0.23\textwidth]{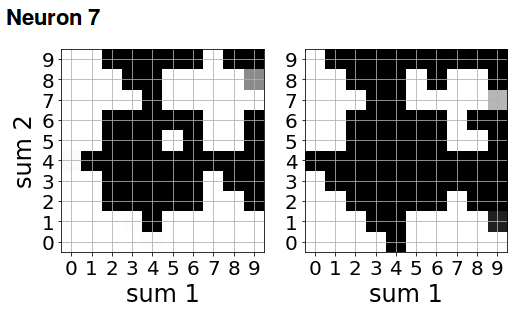}\\
    \includegraphics[width=0.23\textwidth]{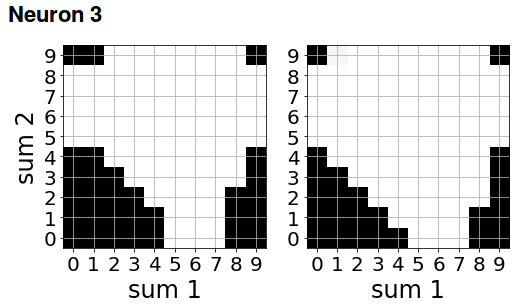}
    \includegraphics[width=0.23\textwidth]{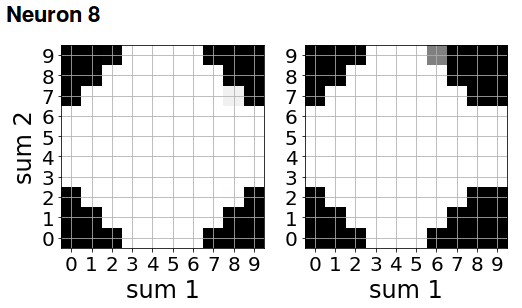}\\
    \includegraphics[width=0.23\textwidth]{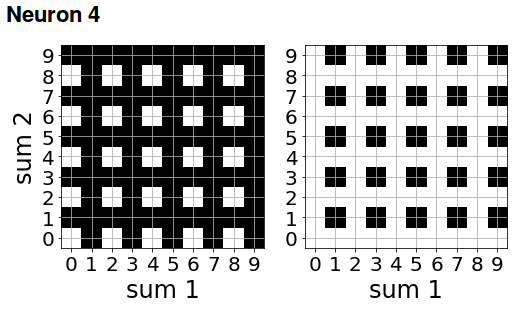}
    \includegraphics[width=0.23\textwidth]{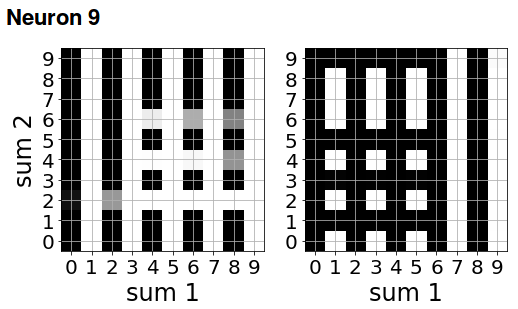}
    \caption{Activation plot of the ten non-regularized neurons in an Elman RNN trained on the addition problem in base 10. For each neuron, the left plot represents the activation when there is no carry, while the right plot represents the activation when there is a carry from the previous step.}
    \label{fig:elman_sum10}
\end{figure}

This is the case for the addition problem when we consider higher bases. Figure  \ref{fig:elman_sum10} shows an example for $B=10$. The network has $20$ recurrent units and has been trained with the previous set of parameters. No errors are observed on the test dataset after training. As expected, only a few units survive the regularization and their activation is binary. However there is now only one neuron (N1) with a straightforward interpretation, and not surprisingly it is coding the carry. The other active neurons need to be used to compute the network's output and, although their activation forms some characteristic patterns, their behavior is not meaningful at all. The extracted automaton is not shown because of its huge complexity.

\begin{figure}[h]
	\centering
	\includegraphics[width=0.23\textwidth]{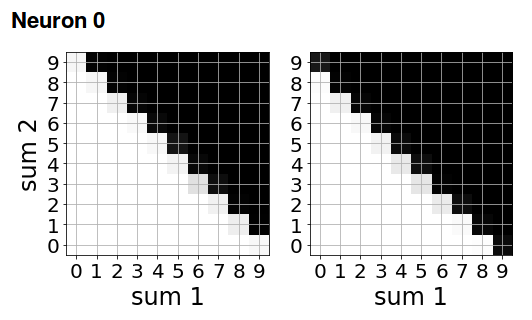}
	\includegraphics[width=0.23\textwidth]{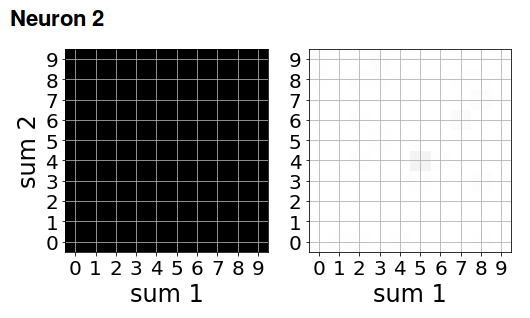}
	\caption{Activation plot of the two non-regularized neurons in a Dual RNN trained on the addition problem in base 10. For each neuron, the left plot represents the activation when there is no carry, while the right plot represents the activation when there is a carry from the previous step.}
	\label{fig:dual_sum10}
\end{figure}

A Dual network, with $10$ neurons in the feedforward layer, has also been trained in the same conditions. After training, all the general observations extracted for the Elman case are still valid, but now only two neurons are active in the recurrent layer (figure \ref{fig:dual_sum10}). These two neurons are necessary to deal with the carry, and this is the main difference with respect to the Elman network: now the recurrent layer memorizes only what is strictly necessary to cope with time dependencies. The carry information stored in the recurrent layer, together with the network input, is sufficient for the additional feedforward layer to compute the correct output. This observation is valid regardless of the numerical base\footnote{We have tested with different $B$ values and in all the cases we obtain the same solution.}.  

\begin{figure}[h]
    \centering
    \includegraphics[width=0.25\textwidth]{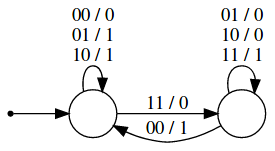}
    \caption{Minimum DFA extracted from a Dual RNN trained on the addition problem in base 2.}
    \label{fig:suma_mealy}
\end{figure}

The corresponding automaton is shown in figure \ref{fig:suma_mealy}. The result for the network trained with $B=2$ has been used for the sake of clarity, but networks trained with different bases provide the same transition diagram (with additional labels for different input/output pairs). The automaton extracted from the Dual network is again a Mealy machine, but now the advantage over the Moore version associated to the Elman RNN is more evident (compare with the DFA shown in figure \ref{fig:suma_moore}). As $B$ increases, the difference between the number of states needed by the recurrent layer in the two network architectures becomes more dramatic. While the Elman network needs more and more additional recurrent units to code the solution, the Dual RNN uses always the same memory configuration, leaving the main part of the computation to the feedforward layer. In summary, by allowing some of the processing be carried out by the feedforward layer, the Dual RNN discharges much of the computational load from the recurrent layer, letting it concentrate on just the information that must be remembered for future time steps. This is a very exciting result that could be of general application in more complex problems.  

\subsection{Arithmetic expressions}
\label{subsec:arithmetic-expressions}

The last experiment tests the network capacity to generate arithmetic expressions with fixed parentheses depth. We consider Elman RNNs with 10, 20 and 30 units, Dual RNNs with 5, 10 and 15 units in the recurrent layer and 10 and 20 units in the feedforward layer, and LSTM networks with 10, 20 and 30 units. We train 10 different networks for each configuration and select the one with the lowest cross-entropy loss on the test set. The selected networks are then used to generate expressions, and evaluated according to the accuracy measure defined in section \ref{subsec:experiments_expressions}. The whole process is repeated 10 times in order to obtain average results. In all the cases we use a batch size of 20, with the rest of parameters as in previous sections.

The results obtained for the Elman RNN are shown in table \ref{tab::results_precision_elman}. The first half of the table corresponds to vanilla Elman RNNs (no noise) with and without L1 regularization. The second half contains the results for noisy Elman RNNs, as described in section \ref{subsec:elmanRNN}, also in the two regularization conditions. We observe that both the noise and the regularization contribute benefits when considered in isolation. Noise introduction allows the networks reach much higher accuracy (compare for example the values 91.8$\pm$12.8 and 70.5$\pm$16.2 for the $10$ units case). The results when L1 regularization is applied show a more modest increase in accuracy, but on the other hand present a much lower variance, which is always desirable. The combination of both noise and regularization does not improve these results. Nevertheless, we must take into account that the networks used in previous sections have been used here with no systematic search for the hyperparameters. Fine tuning may be necessary to adapt the networks to this new problem. 

\begin{table}
	\centering
	\caption{Average accuracy of different Elman RNNs trained to generate arithmetic expressions.}
	\label{tab::results_precision_elman}
	\begin{tabular}{lclrr }
		\toprule
		\textbf{Config} & \textbf{Units} & \textbf{Test} & \textbf{Min} & \textbf{Max} \\
		\midrule
		\textit{noise = 0.0}   & 10 & 70.5$\pm$16.2 & 51.2 & 97.3 \\
		\textit{L1 = 0.0}      & 20 & 79.0$\pm$23.8 & 39.1 & 99.0 \\
	                           & 30 & 41.7$\pm$7.5 & 28.9 & 57.3 \\
		\midrule
		\textit{noise = 0.0}   & 10 & 85.6$\pm$3.1 & 80.0 & 92.0 \\
		\textit{L1 = 0.1}      & 20 & 88.2$\pm$2.2 & 83.2 & 91.3 \\
		                       & 30 & 87.4$\pm$3.2 & 82.0 & 92.6 \\
		\midrule
		\textit{noise = 1.0}   & 10 & 91.8$\pm$12.8 & 60.6 & 99.0 \\
		\textit{L1 = 0.0}      & 20 & 89.9$\pm$9.9 & 67.1 & 98.9 \\
		                       & 30 & 51.2$\pm$24.9 & 22.4 & 98.8 \\
		\midrule
		\textit{noise = 1.0}   & 10 & 80.7$\pm$5.0 & 71.0 & 88.2 \\
		\textit{L1 = 0.1}      & 20 & 82.8$\pm$6.3 & 75.2 & 99.2 \\
		                       & 30 & 60.7$\pm$13.7 & 47.0 & 85.2 \\
		\bottomrule
	\end{tabular}
\end{table}

\begin{figure*}[t]
    \centering
    \includegraphics[width=0.9\textwidth]{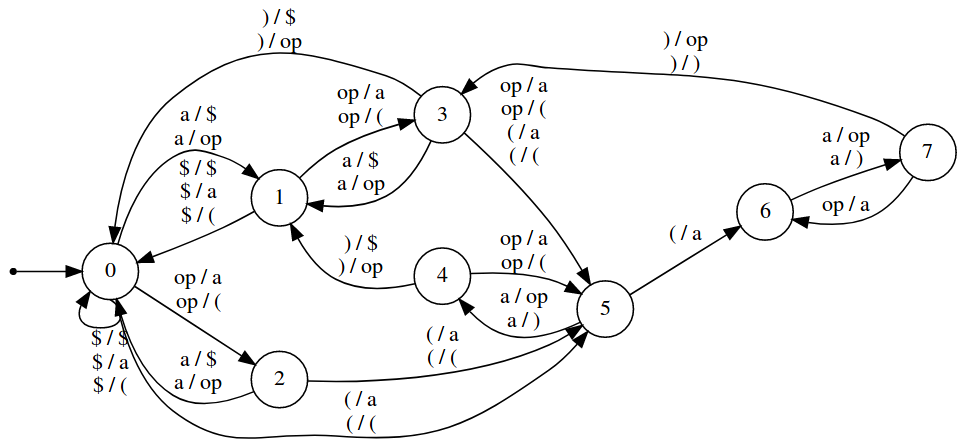}
    \caption{Minimum DFA extracted from a Dual RNN trained on the arithmetic expressions problem. Only the states with a parenthesis depth lower than or equal to 2 are plotted in order to simplify the graph. Transitions which are produced with a probability lower than $0.001$ are also omitted.}
    \label{fig:automata_expAlg_prof2}
\end{figure*}

The Dual RNN provides similar results when no noise is applied (not shown). When noise is injected, however, the results considerably improve (see table \ref{tab::results_precision_dual}). With no regularization, all the considered networks reach more than 99\% accuracy on the generation task. When L1 regularization is included, the accuracy decreases but it is still higher than for the Elman network. In all the cases the recurrent units get binarized and the activation patterns in the recurrent layer may be used to extract a DFA. As for previous problems, this automaton has the form of a Mealy machine. As an example, figure \ref{fig:automata_expAlg_prof2} shows the automaton extracted for the non-regularized 5-10 Dual network. In order to simplify the transitions graph, only the states with a parentheses depth lower than or equal to $2$ are plotted and only transitions with a probability higher than $0.001$ are represented. It is worth noting that these low probability transitions are  responsible for the few grammatical errors observed in the generated strings, hence by discarding them we obtain an automaton that consistently generates the grammar.

\begin{table}
	\centering
	\caption{Average accuracy of different Dual RNNs trained to generate arithmetic expressions.}
	\label{tab::results_precision_dual}
	\begin{tabular}{lclrr }
		\toprule
		\textbf{Config} & \textbf{Units} & \textbf{Test} & \textbf{Min} & \textbf{Max} \\
		\midrule
		\textit{L1 = 0.0}    & 5 - 10   & 99.2$\pm$0.6 & 97.9 & 99.8 \\
		                     & 5 - 20   & 99.3$\pm$0.5 & 98.1 & 99.9 \\
							 & 10 - 10  & 99.3$\pm$0.5 & 97.9 & 99.8 \\
							 & 10 - 20  & 99.3$\pm$0.6 & 97.9 & 100.0 \\
							 & 15 - 10  & 99.6$\pm$0.3 & 99.0 & 99.9 \\
							 & 15 - 20  & 99.7$\pm$0.1 & 99.5 & 99.8 \\
		\midrule
		\textit{L1 = 0.1}    & 5 - 10   & 86.5$\pm$4.9 & 75.1 & 94.8 \\
		                     & 5 - 20   & 89.5$\pm$3.9 & 82.8 & 96.6 \\
							 & 10 - 10  & 89.5$\pm$3.0 & 85.1 & 96.4 \\
							 & 10 - 20  & 90.3$\pm$2.7 & 83.6 & 92.5 \\
							 & 15 - 10  & 93.0$\pm$2.8 & 89.4 & 97.4 \\
							 & 15 - 20  & 87.7$\pm$4.6 & 79.8 & 96.1 \\
		\bottomrule
	\end{tabular}
\end{table}

Finally, to put these results in a proper context, we also consider the use of LSTMs on the same problem. The results are shown in table \ref{tab::results_precision_lstm}. The main observation is that, perhaps surprisingly, LSTMs are not able to achieve as high accuracy as the Dual networks on this prediction task. The best configuration provides an accuracy that is $3$ percentual points below the value obtained by the Dual RNNs. Although a proper parameter tuning might help improve these results, the use of the proposed Dual network has the additional advantage of a better interpretability derived from activity binarization. While the Dual networks can be easily understood as a DFA in Mealy form, the mechanisms used by LSTMs to solve the problem are not clear at all. On a different plane, the computational resources needed to train a Dual RNN are also lower in both memory and time. 

\begin{table}
	\centering
	\caption{Average accuracy of different LSTM networks trained to generate arithmetic expressions.}
	\label{tab::results_precision_lstm}
	\begin{tabular}{lclrr}
		\toprule
		\textbf{Config} & \textbf{Units} & \textbf{Test} & \textbf{Min} & \textbf{Max} \\
		\midrule
		\textit{L1 = 0.0}  & 10  & 83.7$\pm$3.6 & 79.4 & 88.0 \\
		                   & 20  & 88.7$\pm$1.1 & 87.0 & 90.1 \\
		                   & 30  & 96.5$\pm$0.3 & 96.0 & 96.9 \\
		\midrule
		\textit{L1 = 0.1}  & 10  & 47.3$\pm$1.3 & 45.3 & 49.2 \\
		                   & 20  & 60.1$\pm$0.9 & 58.9 & 61.4 \\
		                   & 30  & 64.9$\pm$1.0 & 63.4 & 66.1 \\ 
		\bottomrule
	\end{tabular}
\end{table}

\section{Conclusions}
\label{sec:conclusion}

In this article we have explored a neural network architecture that combines a recurrent layer connected to the network input and a feedforward layer that processes both the input and the output of the recurrent layer. This way, two different processing paths may focus on different aspects of learning. The recurrent path concentrates its resources on remembering information that must be preserved along time, so working as a kind of memory. The feedforward path is able to use this memory, together with the input, to provide the final network output. Networks using this architecture seem to make a better use of their computational resources, providing better results than traditional Elman RNNs and LSTM networks on prediction problems of different complexity. Additionally, the trained networks are more interpretable. 

We have also observed that the introduction of noise in the activation function of the recurrent units forces these units to behave in a binary manner, with their output being always either $+1$ or $-1$. As a consequence, the time evolution of the network when it is presented a given input sequence can be seen as a transition through a finite set of discrete states. It is then possible to extract a transition map and show that the networks are internally behaving as deterministic finite automata. Although this observation is true also for the Elman RNNs, the automata extracted from networks that use the Dual architecture are much simpler, and implement a correct Mealy machine in all the problems we considered.  

When facing a language generation problem, the Dual RNN also outperforms the Elman and LSTM architectures in terms of the grammatical correctness of the generated expressions, and the interpretability of the network as a Mealy machine is still preserved. In brief, we are able to train better networks that are in addition more interpretable. In spite of the simplicity of the considered language, we expect that this behavior can be extended to more complex problems, with potential applications in fields such as automatic music composition, natural language processing or machine translation. Even if the results on these areas did not achieve state of the art performance, the gain in interpretability might be worth the price. 

\begin{acks}
This work has been partially funded by grant S2017/BMD-3688 from Comunidad de Madrid and by Spanish project MINECO/FEDER
TIN2017-84452-R (\url{http://www.mineco.gob.es/}).
\end{acks}

\bibliographystyle{ACM-Reference-Format}
\bibliography{sigkdd2020}

\end{document}